\documentclass[runningheads]{llncs}
\usepackage{graphicx}
\usepackage{amsmath} %
\usepackage{multirow}
\begin{document}
\title{Resonant Inductive Coupling Power Transfer for Mid-Sized Inspection Robot}
%
%
\author{Mohd Norhakim Bin Hassan\inst{1}\orcidID{0000-0001-5522-2952} \and
Simon Watson\inst{2,3}\orcidID{0000-0001-9783-0147} \and
Cheng Zhang\inst{3}\orcidID{0000-0002-4547-8329}}
\authorrunning{M. Bin Hassan et al.}
%
\institute{
Department of Electrical and Electronic Engineering, University of Manchester, Manchester M13 9PL, United Kingdom 
\email{\{mohdnorhakim.binhassan, simon.watson, cheng.zhang\}@manchester.ac.uk}}
\maketitle              
\begin{abstract}
This paper presents a wireless power transfer (WPT) for a mid-sized inspection mobile robot. The objective is to transmit 100 W of power over 1 meter of distance, achieved through lightweight Litz wire coils weighing 320 g held together with a coil structure of 3.54 kg. The Wireless Power Transfer System (WPTS) is mounted onto an unmanned ground vehicle (UGV). The study addresses an investigation of coil design, accounting for misalignment and tolerance issues in resonance-coupled coils. In experimental validation, the system effectively transmits 109.7 W of power over a 1-meter distance, with obstacles present. This achievement yields a system efficiency of 47.14\%, a value that is remarkably close to the maximum power transfer point (50\%) when the WPTS utilises the full voltage allowance of the capacitor. The paper shows the WPTS charging speed of 5 minutes for 12 V, 0.8 Ah lead acid batteries. 


\end{abstract}
\section{Introduction}

The use of robots is on the rise, and as tasks increasingly require mobility beyond the constraints of a fixed workspace, there is a growing demand for deploying mobile robots \cite{b1}. This can be observed from the vast development of various types of mobile robots at an unprecedented pace of operation in the ground, air, and underwater. Several implementations for such mobile robots include oil and gas refinery inspection \cite{b2}, radiation mapping \cite{b3} \cite{b4}, underwater mapping \cite{b5} \cite{b6}, and nuclear-decommissioning \cite{b7} \cite{b8}.

The main constraint for long-term mobile robot deployment is the onboard battery capacity, especially in scenarios where robots operate behind sturdy concrete walls \cite{b1}. Whilst current technological advancements have increased the energy capacity and output power of batteries, this capacity is still insufficient for mobile robots. As a result, the only options are to return to the charging station for recharging or to perform manual battery replacement on-site \cite{b1}. These two options merely address the mobile robots' battery capacity limitation, albeit at the expense of increased downtime. 

The other option is to use a tether to control the mobile robot. However, this approach has additional difficulties such as tether crossover, restrictions on bending around obstacles, and a decrease in the payload of the mobile robot from the tether system, all of which restrict the mobile robots' performance and mobility \cite{b9}. One feasible solution to the battery capacity limitation for mobile robots is the use of wireless power transfer (WPT) technologies.

The common implementations of WPT are mainly concentrated on either high-power applications (kW) such as electrical vehicles which generally operate with a transmission distance of less than 0.3 m \cite{b12} or low-power (less than a few Watts) applications such as consumer electronic devices and medical implants within centimetre distances and wireless sensor network (WSN) at kilometre distances \cite{b11}. Cheah et al. discussed WPT technologies limitations and the implementation for mobile robots with mid-power range and transmission distances of from 1 m up to 20 m. 

This paper discusses the implementation of WPT technology into mobile robot applications. The aim is to incorporate WPT effectively while maintaining robustness without contributing a significant weight to the robot's payload. The use of WPT in this paper is limited to a mid-sized inspection robot with a nominal operating power of 100 W and transmission distances of up to 1 m. The objective is to reach a 50\% efficiency target at the maximum power point, thus fully leveraging the stress-handling capabilities of the involved components. The selection of transmission distance is based on the deployment of mobile robots in remote areas, potentially encountering gaps (whether air or obstacles).

\section{Mobile Robot \& Wireless Power Transfer Systems}
\subsection{Mobile Robot}
A mid-sized Unmanned Ground Vehicle (UGV) is taken into consideration to be a testing platform at the initial stage of conceptual approval apart from the other various types of mobile robots. Hence, a UGV Agile X Scout Mini is chosen. It is a compact entry-level field mobile robot for the research platform for research and development purposes with a four-wheel differential drive and independent suspension. The platform weighs 26 kg and has a maximum permissible payload of 10 kg. Scout Mini has been chosen to provide insight into the practical implementation aspects concerning the proportional relationship between the coil frame's size and that of the mobile robot. The problem statement revolves around assessing the viability of a WPT technology to supply necessary power across various transmission mediums for a moving mobile robot operating within mid-range distance. This study exclusively concentrates on an end-to-end WPT system, wherein power is directly transmitted to the robot's receiver without the need for intermediary relay systems.

\subsection{Wireless Power Transfer Technologies}

WPT is an umbrella term for a variety of technologies that use electromagnetic fields to transmit energy. The transmitting distance over which the system can efficiently transfer power differs between these technologies. This is determined by whether the energy-transmitting means are directional or non-directional. Instead of using conventional cables, WPT technology can transmit energy from the power source to the target through non-conducting mediums such as air, concrete or water \cite{b13}. 

There are two primary categories of WPT which are radiative and non-radiative. Within the realm of radiative WPT, two prominent techniques are microwave and laser. On the other hand, examples of non-radiative WPT include several technologies, including inductive coupling, magnetic resonance inductive coupling and capacitive coupling. Each of these technologies can be further classified into either direct or background energy harvesting. The distinction lies in their energy source: direct energy harvesting retrieves power from a specifically established transmitter. Meanwhile, background energy harvesting utilizes ambient energy incidental to other processes, like heat from a cooling system or radio waves from wireless communication.

Far-field WPT is usually used to accomplish a longer range of power transmission. This often includes an extended distance to be covered in one transmission. Laser beams and microwaves are the two best-suited forms of electromagnetic radiation techniques for power transmission. The transmission path for a high-efficiency microwave system must be unbroken, and it must be extremely directed \cite{b23}.

A near-field magnetic solution will enable transmission through numerous barriers, including walls, obstructions, and even humans. The magnetic field remains unaffected by these factors, offering superior tolerance when compared to the far-field WPT. The near-field technique makes use of the inductive coupling effect of non-radiative electromagnetic fields, including the inductive and capacitive mechanisms \cite{b24}. 

\subsection{Analysis}
Tab. \ref{tab: wpt_summary} summarised the main characteristics of different WPT technologies through non-conducting mediums. Parameters of interest are the transmission range ($T_r$), transmitter-to-receiver diameter ratio ($R_{Tx-Rx}$) and the maximum efficiency ($\eta_{max}$). Laser and microwave WPT can be eliminated due to higher potential hazardous effects on humans within its radiative field and line-of-sight requirement which requires an accurate tracking system. The Maximum Permissible Exposure (MPE) can be determined using IEEE C95.1-2005 regulations \cite{b25}. Capacitive coupling is constrained to short-range operations, limited to a maximum transmission distance of 0.3 m. This restriction comes from the minimal capacitance generated by the permittivity of space, leading to a small transmitter-to-receiver ratio. Enhancing capacitance involves enlarging capacitive plate dimensions and reducing transmission distance, but these adjustments pose challenges given the limited dimensions of the receiver on a mobile robot.

\begin{table}
\begin{center}
        \centering
        \caption{Summary of different performances of WPT technology based on the range of transmission distance with respect to transmitter-to-receiver ratio \cite{b1}.}
        \begin{tabular}{|c|c|c|c|}
\hline 
WPT Technology & $T_r$ & $R_{Tx-Rx}$ & $\eta_{max}$   \\
\hline
Laser & \multirow{2}*{$d$ $\ge$ 20 m} & 916 \cite{b14} & 14\% \cite{b15}  \\

                           Microwaves &&  222 \cite{b16} & 62\% \cite{b17}  \\

\hline
Capacitive & 0.1 m $\leq$ $d$ $\leq$ 0.3 m      & 0.5 \cite{b18} & 90\% \cite{b18}\\
                           Inductive &  0.1 m $\leq$ $d$ $\leq$ 20 m  & 3.5 \cite{b19}  & 98\% \cite{b20}  \\
                           Resonance Inductive & 0.1 m $\leq$ $d$ $\leq$ 5 m  & 6 \cite{b21} & 85\% \cite{b22} \\
\hline
\end{tabular}
\label{tab: wpt_summary}
\end{center}
\end{table}

The inductive coupling power transfer is accomplished by employing an elongated ferrite core to ensure magnetic flux lines connect between the transmitter and receiver. Nonetheless, the transmitter utilized is substantial and inconvenient, leading to an increased transmitter-to-receiver ratio. While inductive coupling has demonstrated a higher transmission range compared to resonance inductive coupling, the transmitter-to-receiver ratio is lower, suggesting that resonance inductive coupling can indeed achieve a greater transmission range. As discussed in the previous subsections, a combination of a UGV which is Scout Mini and the resonance inductive coupling power transfer are the most feasible for the context of this paper.

\subsection{Parameters of Resonant Inductive Power Transfer System} 

Resonant Inductive Power Transfer (RIPT) is regarded as a unique instance of inductive coupling power transfer, where strong electromagnetic coupling is attained by operating at the resonance frequency of the coils. This operational principle can be realized using two or more coils. Two operational principles exist; power delivered to load (PDL) and power transfer efficiency (PTE), where a trade-off between power delivered and system efficiency is observed. High system efficiency is influenced by the subsystems such as the power supply, coil configuration, and coil material.

\begin{figure}
\centering\includegraphics[width=\textwidth]{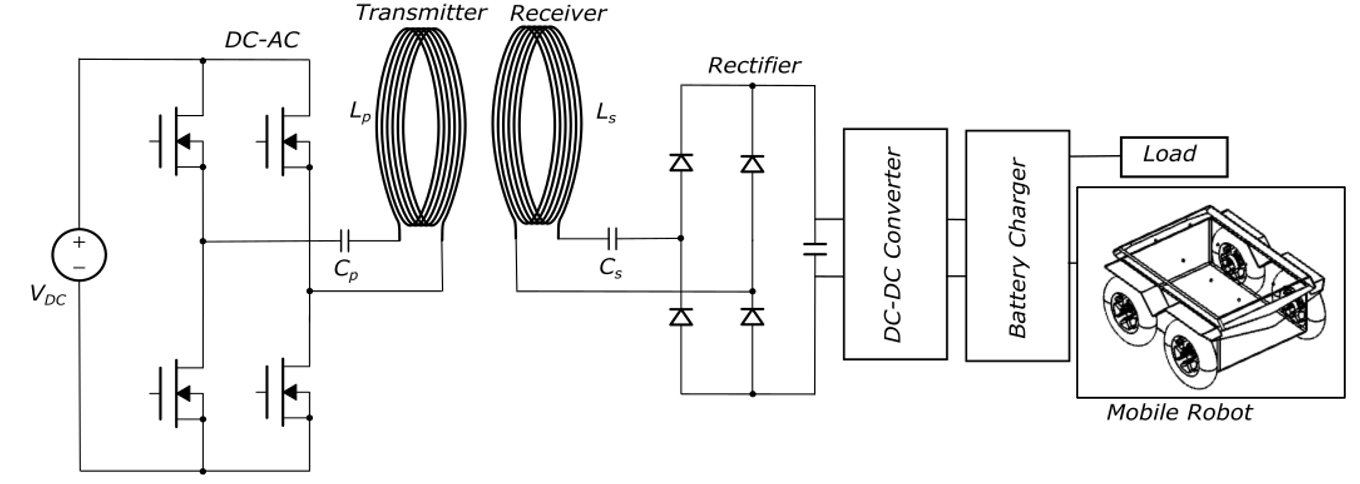}
\caption{Circuit topology of resonant inductive coupling for a mid-sized inspection mobile robot.} \label{fig_ipt2}
\end{figure}

Fig. \ref{fig_ipt2} depicts the simplified circuit topology of the WPT for a mid-sized inspection mobile robot. Equation \ref{eq: eq1} shows the relationship of angular frequency between self-inductance and compensation capacitors at the primary and secondary coils \cite{b17}. Ideally, the system operates at the nominal frequency ($f_s$) where both sides of the resonators shall be tuned as in eq. \ref{eq: eq1} where $L_p$ and $L_s$ are the self-inductance, while $C_p$ and $C_s$ are the compensation capacitors at the primary and secondary sides respectively. 

\begin{equation}
\label{eq: eq1} 
f_s=\frac{1}{2 \pi\sqrt{L_{p}C_{p}}}=\frac{1}{2 \pi\sqrt{L_{s}C_{s}}}  
\end{equation}

The efficiency of energy transmission is primarily influenced by the load impedance, along with other operational factors including frequency, tolerances of inductive and capacitive components, and alignment. Eq. \ref{eq: eq2} can be used to obtain the coupling coefficient ($k$) from the self- and mutual inductances ($M$):

\begin{equation}
\label{eq: eq2} 
k=\displaystyle \frac{M}{\sqrt{L_{p}L_{s}}}, 
\end{equation}
    
By employing rms values for voltages and currents, the output power across the load ($P_{R_L}$) can be determined using the expression where $V_L$ is the load voltage, $R_L$ is the resistance across the load, and $I_s$ is the ac secondary current:

\begin{equation}
\label{eq:load_power_org}
P_{R_L}=\frac{\left|V_L\right|^2}{R_L}=\left|I_s\right|^2R_L 
\end{equation}

$I_p$ and $I_s$ can be expressed using matrix calculation in eq. \ref{resonance_matrix} using the component of Z-matrix in eq. \ref{z_matrix}, where $R_p$ denotes the resistance on the primary side, $R_s$ the resistance on the secondary side, and $V_p$ the external source voltage.

\begin{equation}
\begin{bmatrix}
\label{z_matrix}
  Z
\end{bmatrix}=
\begin{bmatrix}
R_p + j\omega L_p + \frac{1}{j\omega C_p}  & j\omega M \\
j\omega M & R_s + R_L + j\omega L_s + \frac{1}{j\omega C_s}
\end{bmatrix} 
\end{equation}

The relationship between $Z_{1_2}$ and $Z_{2_1}$ can be expressed as $Z_{1_2}$=$Z_{2_1}$. During resonance condition, it simplifies as: 

\begin{equation}
\begin{bmatrix}
\label{resonance_matrix}
R_p & j\omega M \\
j\omega M & R_s+R_L
\end{bmatrix}
\begin{bmatrix}
I_p \\
I_s
\end{bmatrix}=
\begin{bmatrix}
V_p \\ 
0
\end{bmatrix} 
\end{equation}

The current at the primary side of the coil ($I_p$) is given in eq. \ref{eq:Ip}  whilst $V_p$ is given in eq. \ref{eq:Rp} as the following:

\begin{equation}
\label{eq:Ip}
I_p = \frac{j(R_s+R_L)\cdot I_s}{\omega M} 
\end{equation}

\begin{equation}
\label{eq:Rp}
R_p\cdot I_p + j\omega M\cdot I_s = V_p
\end{equation}

Substituting eq. \ref{eq:Ip} into eq. \ref{eq:Rp}:

\begin{equation}
\label{eq:Vp}
R_p\cdot\left(\frac{j(R_s+R_L)\cdot I_s}{\omega M}\right)+j\omega M\cdot I_s = V_p
\end{equation}

Hence, $I_s$ is obtained as follows:
\begin{equation}
\label{eq:sec_current}
I_s =-V_p\left(\frac{j\omega M}{R_p(R_s+R_L)+(\omega M)^2}\right) 
\end{equation}

Substituting eq. \ref{eq:sec_current} into eq. \ref{eq:load_power_org} to obtain power across the load:

\begin{equation}
\label{eq:P_load}
P_{R_{L}}=\frac{(\omega M)^2V_p^2R_L}{\left(R_p(R_s+R_L)+(\omega M)^2\right)^2} 
\end{equation}

Whilst the power transfer efficiency, defined as the ratio of the output power across the load $P_{R_L}$ to the input power delivered to the primary coil ($P_{in,AC}$), can be expressed as follows:
    
\begin{equation}
\label{eq:eff}
\eta=\frac{P_{R_{L}}}{P_{in,AC}}=\frac{(\omega M)^{2}R_{L}}{(R_{s}+R_{L})(R_{p}(R_{s}+R_{L})+(\omega M)^{2})} 
\end{equation}

The self and mutual inductances of the coils and their equivalent series resistances have been discovered to have a direct impact on coupling effectiveness and power transfer efficiency. 

\section{Constraints} 

\subsection{Dimensional}
The resonator is formed by the connection of external compensation capacitors with each coil. Practically the winding also has its parasitic resistance and capacitance. $X_n$ is the total series reactance of resonator circuit $n$ and $R_n$ is the total equivalent series resistance. The transmitter circuits are connected to a voltage or current source and the receiver circuits are connected to loads \cite{b26}.

The mutual and self-inductance values are determined by the physical dimensions and positions of the winding, unlike other values which are appointed externally. The inductance values can be calculated from the vectors of winding segments using Neumann's formula \cite{b27}.

\begin{equation}
M_{i_j} = \frac{\mu_0}{4\pi} \oint\oint \frac{\mathrm{d}l_i\cdot\mathrm{d}l_j} {\mid{r\mathrm{d}l_i\cdot r\mathrm{d}l_j}\mid} 
\end{equation}

where $l_i$ and $l_j$ are contours of two coils, $\mathrm{d}l_i$ and $\mathrm{d}l_j$ are infinitesimal segment vectors on the two coils and $\mathrm{r}$ vectors are the displacement vectors refer to the reference origin. In the case that $i$ = $j$, $M_{i_j}$ becomes the self-inductance of the coil and the integral term cannot be computed if $\mathrm{d}l_i$ and $\mathrm{d}l_j$ are the same segments. An alternative closed-form formula is used to replace the integral term. For wires with a circular cross-section, the following formula can be used \cite{b28}.

\begin{equation}
L_{part} = \frac{\mu_0}{2\pi} 
\Bigg[
l\,log \Bigg (\frac {l+\sqrt {l^2+\rho^2}}{\rho} \Bigg) - \sqrt{l^2+\rho^2} + \frac{l}{4} + \rho
\Bigg] 
\end{equation}

where the cross-section's radius is $\rho$ and the segment's length is $l$. The predicted self-inductance is the best predictor of the measurement value when the segment's length is equal to the wire's diameter.

\subsection{Power Losses, Interference and Translational Offsets}

Additional elements like fine-tuning coil designs, effectively managing heat, implementing shielding, and mitigating electromagnetic interference (EMI) ultimately play a pivotal role in assessing the efficiency of WPT, depending on the specific approaches employed. Higher transmission frequencies (f $\geq$ 50 kHz), resonant switching, and optimised coil components like HF-litz wire and ferrite cores can all help achieve this \cite{b29}. 

Special consideration is needed for the interactions between transmission and reception information and the AC charging flux in electronic circuit loads. The high charging flux density can induce eddy currents in unintended metallic components within these loads, leading to internal temperature increases and circuit damage. Additionally, power losses occur in the secondary and primary circuits, coils, and magnetics. Unusual entities like ferromagnetic or metallic materials near the flux routes can also absorb radiated power. When such materials are positioned in the AC magnetic flux, they generate induced eddy currents, resulting in temperature rise and conduction losses. Significant conduction loss may lead to safety concerns and potential system damage or failure \cite{b30}. For instance, a power loss of 0.5–1 W in metallic materials can elevate their temperature above 80\textdegree C \cite{b31}.

RIPT typically involves tuning the primary and secondary coils to the same resonant frequency, facilitating efficient power transfer across a specific distance. However, achieving perfect alignment between the coils may be impractical in real-world scenarios. Translational offsets arise from this scenario, enabling the coils to transfer power effectively even when not perfectly aligned, albeit possibly with reduced efficiency. It presents both advantages and challenges. On one hand, it permits power transfer across various orientations and distances, thereby enhancing system flexibility and robustness. On the other hand, it complicates system design and optimization, as engineers need to consider factors like coil geometry, alignment tolerances, and potential interference from nearby objects \cite{b32}. This fits perfectly with the implementation of RIPT for the mobile robot which provides charging flexibility when movement and space are limited.

\section{Simulations and Experimental Setup}

\begin{figure}[h]
\includegraphics[width=\textwidth]{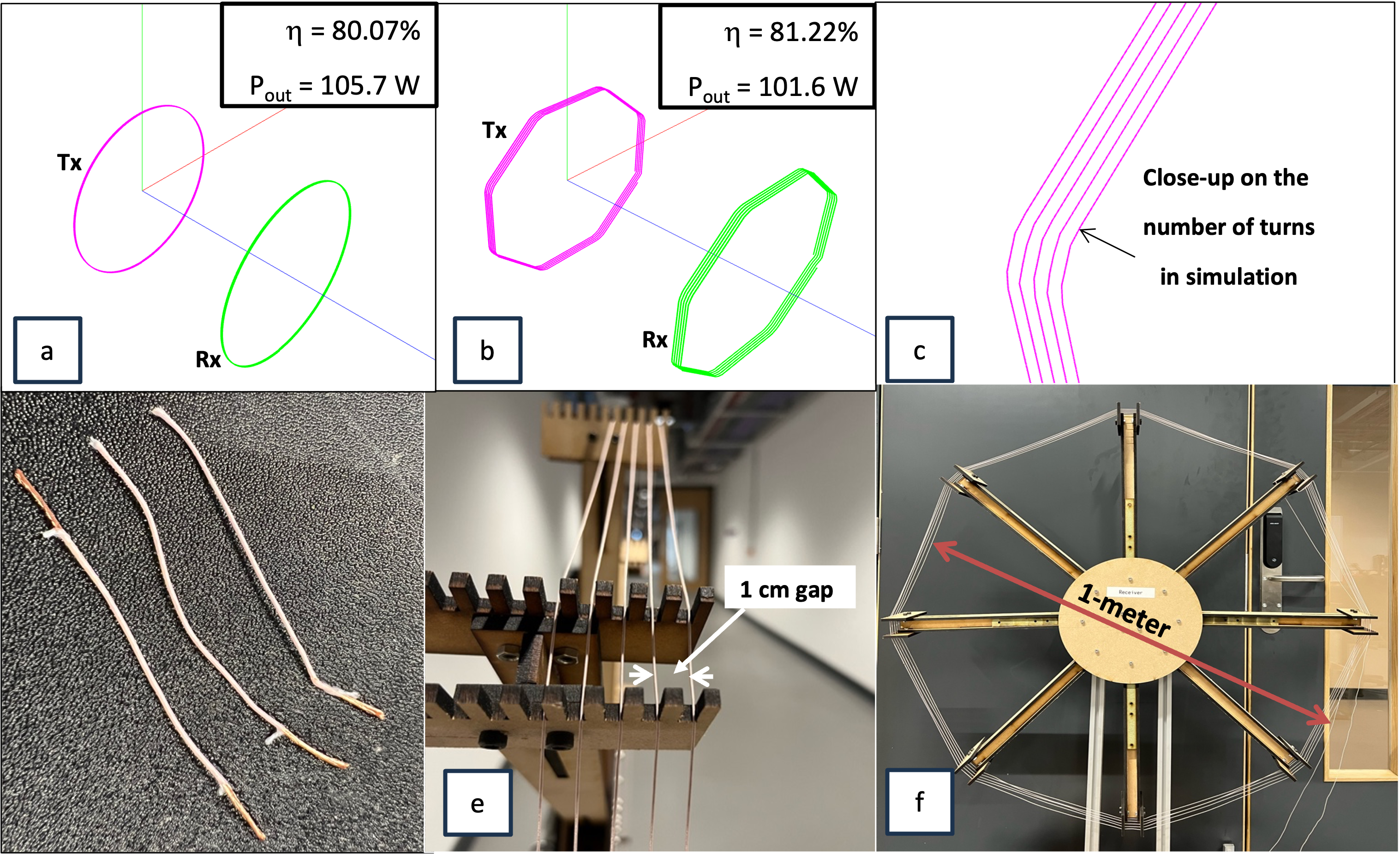}
\caption{a) Simulated circle coils, b) simulated octagon-shaped coils, c) a close-up view of the number of turns for each coil in the simulation,  d) A sample of lightweight litz wire to construct the transmitter and receiver coil, e) a 1 cm gap for each individual turn, f) an aperture of 1 m for the transmitter and receiver coils.}
\label{frontsidecoil}
\end{figure}

The optimization method for designing the coil shape and structure was conducted to identify a collection of solutions that effectively balance multiple competing objectives. In this specific context, there are two key objectives: the quantity of coil turns and the transmitted power. The simulation was conducted using IPTVisual \cite{b27} to model coils based on the parameters of interest. The transmitter (TX) and receiver (RX) coils are depicted in fig. \ref{frontsidecoil}a in magenta (left) and green (right) respectively to show the simulation setup of the circular coils with 0.75 mm cross-sectional radius wires and five-turn-per layer helical structure each. The total length of each coil is 15.363 m. This is used as the base of the simulation for which this will be the specification of the wire aimed to be used in the experimental setup.

The self-inductance for both transmitter and receiver coils at resonance condition evaluated from IPTVisual is 63.15 $\mu H$ ($L_p = L_s$) and is compensated with a capacitance of 1 nF at each side. A 10 $\Omega$ load is applied to the receiver circuit. Simulation results show that five coil turns are the optimal amount of winding which provides a balance between the coil resistance, coil length, WPT efficiency and the amount of power to be transmitted. For practicality, octagonal frames are employed to secure Litz wire coils. They closely resemble circular coils. Each frame is octagonal with a 1-meter opening, and the coil has five turns of Litz wire.

Fig. \ref{frontsidecoil}a, \ref{frontsidecoil}b and \ref{frontsidecoil}c depict the simulations for two geometrical shapes (circular and octagon coils). Simulation results from the circle diameter of 1 m and inscribed octagon of the same diameter show a close approximation of efficiency for the ideal case (no external disturbances in 3D space) in simulation (80.07\% for circular, 81.22\% for octagon). Both have been simulated with the same transmission distance of 1 m and approximation results of 100 W to match the common operating power of a mid-sized inspection mobile robot (105.7 W for circular, 101.6 W for octagon) with 5 coil turns of identical self-inductance and mutual coupling. The variation in transmitted power between circular and octagonal coils can be attributed to the distinct allocation of coil length when shaping them according to their respective geometries. This disparity in coil length distribution results in differing coil resistances, which, in turn, impacts both the resonance frequency and overall output power. Thus, octagon coils were selected for the experimental setup as the results approached the system efficiency of circular coils and matched the objective of achieving 100 W of transferred power.

Fig. \ref{frontsidecoil}d shows the sample for the lightweight litz wires used for the coils. Fig. \ref{frontsidecoil}e shows the 1 cm gap between each coil turn designed to mitigate proximity effects. Fig. \ref{frontsidecoil}f shows the diameter of the structure frame to hold the litz wire in place forming the transmitter and receiver coils. The experiment will utilise the implementation of RIPT and both coil frames are designed to hold the coil turns through a wall or obstacles of 1 m distance separating these two coils. The observation of power transmission efficiency is expected to achieve a minimum of 50\%.

\begin{figure}[h]
\centerline{\includegraphics[width=\textwidth]{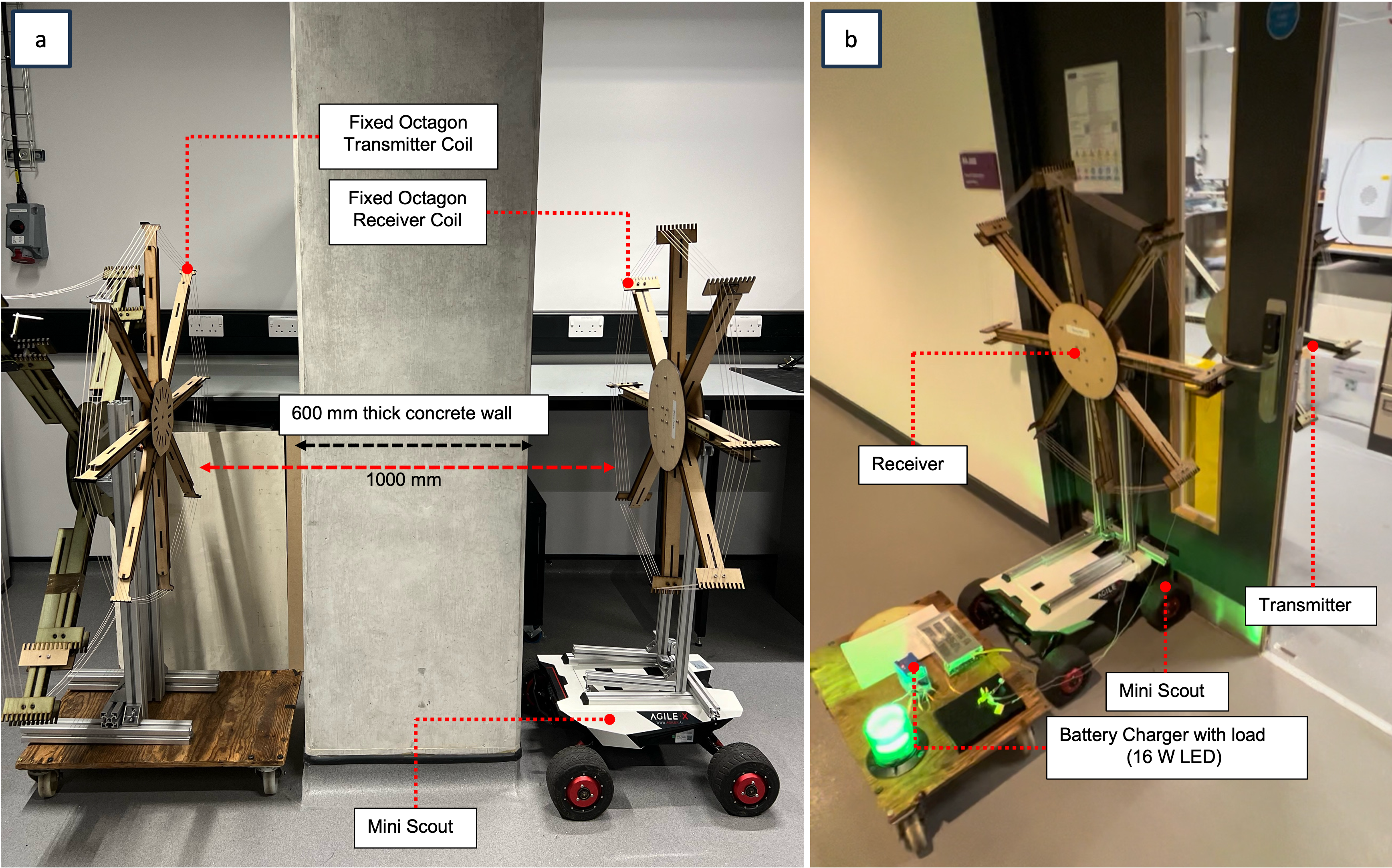}}
\caption{a) Transmitter and receiver octagon coil frames in between 600 m thick concrete wall, b) Powering up 16 W LED with WPT with obstacles in between transmitter and receiver coils.}
\label{scout_16W_LED}
\end{figure}

The transmitter coil frame is separately mounted at a height of approximately 0.9 m from the ground, aligned coaxially with the receiver coil frame. A receiver coil is installed onto the Agile X Mini Scout (refer Fig. \ref{scout_16W_LED}a). The WPT was powered by a GS61004B-EVBCD evaluation board. Table \ref{tab:ipt_param} shows the key parameters of the experimental RIPT. The receiver coil was linked to a rectifier circuit, a DC-DC power converter, a UPS battery charger, and a 16 W LED with a 12 V, 0.8 Ah lead-acid battery. 

\begin{table}[h]
\begin{center}
        \centering
        \caption{Component parameter for octagon coil WPT (OC-WPT)}
    \begin{tabular}{c c c c}
    \hline
     Parameter  & Symbol & Value & Unit\\
     \hline
     DC supply voltage & $V{_{DC}}$ & 43 & V\\
     Operating frequency & $fs$ & 615 & kHz\\
     Capacitance at primary coil & $C_p$ & 1 & nF\\
     Capacitance at secondary coil & $C_s$ & 1 & nF\\
     Self-inductance at primary coil & $L_p$ & 63.15 & $\mathrm{\mu H}$\\
     Self-inductance at secondary coil & $L_s$ & 65.73 & $\mathrm{\mu H}$\\
     Mutual inductance & $M$ & 1.4525 & $\mathrm{\mu H}$\\
     \hline 
    \end{tabular}
    \label{tab:ipt_param}
\end{center}
\end{table}

Experiment 1 was conducted to test WPT's capability to transmit power through a 600 mm concrete pillar over a 1 m distance (as shown in fig. \ref{scout_16W_LED}a. Experiment 2 was conducted aimed to test WPT for power transmission through metallic obstacles at a 1 m distance. A fire door and metallic bin were placed between the transmitter and receiver coils (as shown in Fig. \ref{scout_16W_LED}b). The results for both experiments showed that WPT transmitted 100 W of power without significant power losses.

\section{Results and Observations}
\subsection{Power Transmission}

The measured results are based on the experimental setups as discussed in the previous section, with an input voltage of 43 V and peak input current (7.284 A), the output voltage (33.12 V) measured across a 10 $\Omega$ load resistance and the receiving power of (109.7 W). The system's resonance operating frequency, determined by inductance and capacitance, was operated at 615 kHz. The input power was half of the output transmitted at the receiver with 1 nF 2500 V C0G (NP0) ceramic capacitance at each side. This targets a 50\% energy efficiency for achieving the maximum power transfer point, ensuring the full utilization of component stress capabilities. A current probe was used to monitor the transmitting circuit's current. Meanwhile, the receiver circuit measures output voltage across the load resistance.

\subsection{Effect of Translational Offsets}

\begin{figure}[htpb]
\centerline{\includegraphics[width=1.0\textwidth]{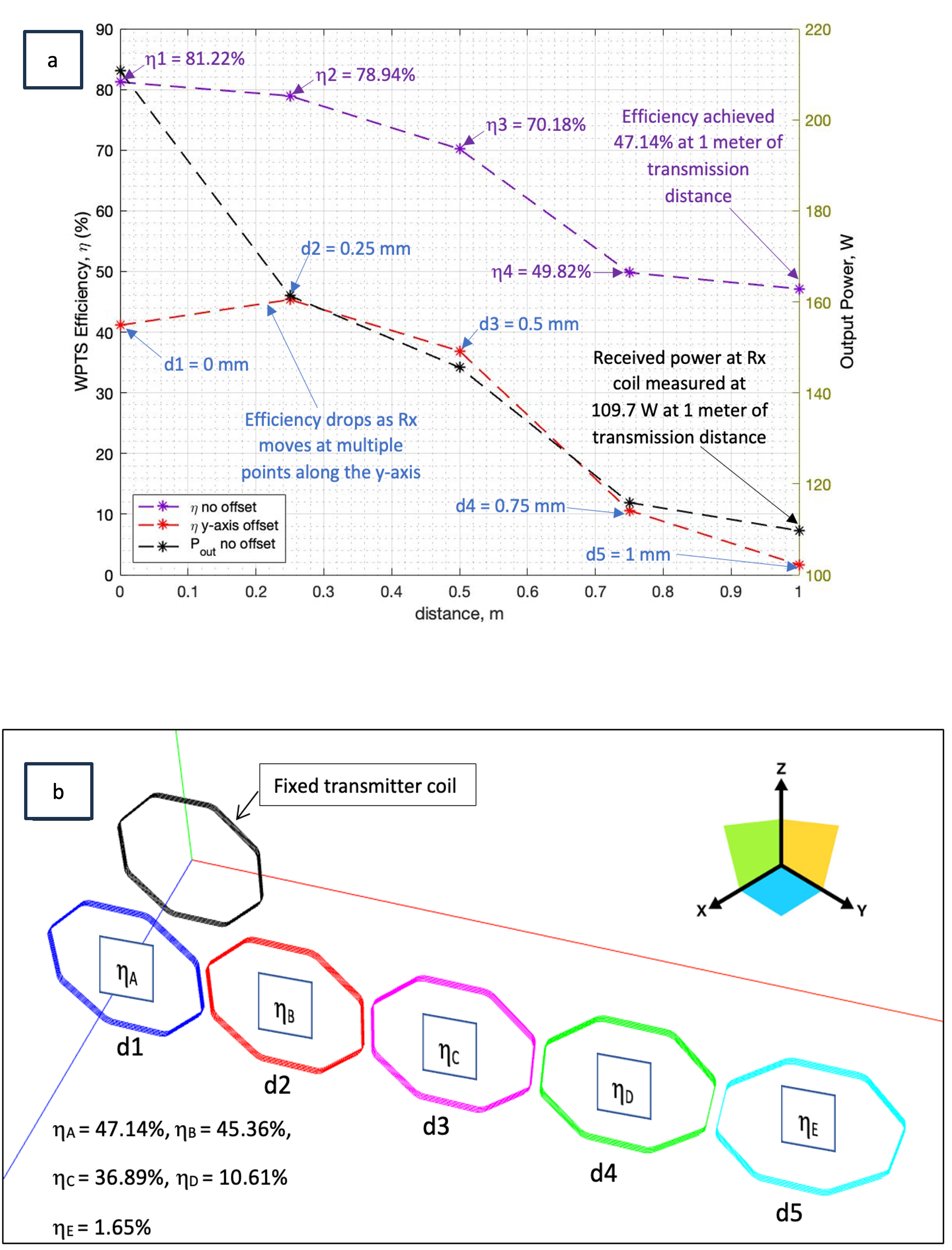}}
\caption{a) WPTS efficiency and output power vs. transmission distance for coaxially aligned coils and efficiency of y-axis offset coils,  b) WPTS efficiency reduces when the receiver coil is positioned at multiple points along y-axis for operational space offset testing (with d5 showing the maximum distance of 1 m)}
\label{eff_offset_yaxis}
\end{figure}

Fig. \ref{eff_offset_yaxis}a illustrates the WPTS efficiency, output power and the transmission distance for coaxially aligned coils and the efficiency of y-axis offset coils. It is observed that the estimated efficiency can be achieved when these two coils are positioned coaxially whilst it reduces when the receiver coil is positioned farther away from the coaxial points. In Fig. \ref{eff_offset_yaxis}b, the offset positions along y-axis proved the reduction of WPTS efficiency with transmitted power as low as 0.466 W when at d5 position. The objective of this experiment is to obtain 100 W receiving power through WPTS and it is proven to be capable with satisfactory efficiency.

\subsection{Battery Charging using WPT}

\begin{figure}[htpb]
\centerline{\includegraphics[width=1.0\textwidth]{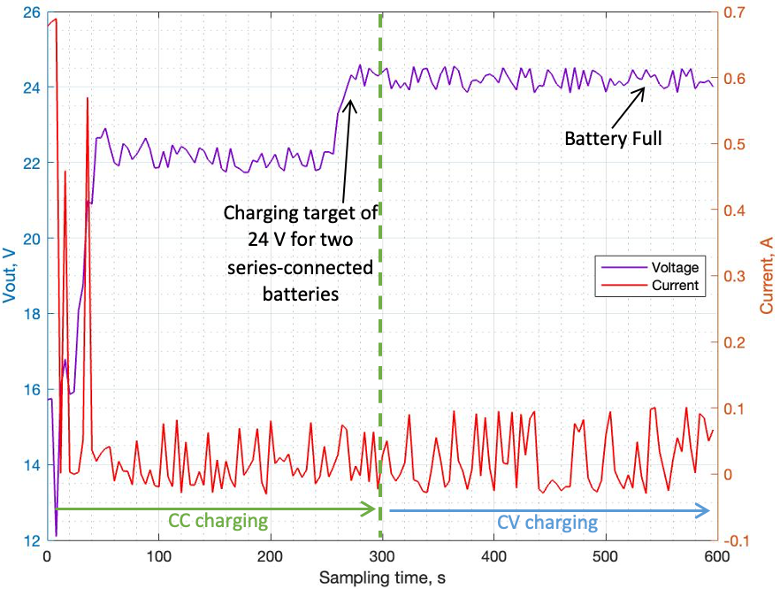}}
\caption{I-V curve of 12 V, 0.8 Ah lead-acid battery charging with 615 kHz using the octagon-shaped transmitter and receiver coil frames.}
\label{batt_charging}
\end{figure}

Smaller 12V, 0.8Ah lead-acid batteries were chosen for a demonstration of the WPTS's battery charging capability for easier observation. Fig. \ref{batt_charging} displays the I-V curve of the lead-acid battery being charged using the WPT. During the charging process, it was observed that the battery charger alternated between charging and discharging modes, as indicated by LED indicators. The voltage input to the battery charger exhibited fluctuations and instability, primarily due to the presence of built-in pulsating diagnostic and maintenance routines for lead-acid batteries. Given the small size of the batteries, these fluctuations were significant. Nonetheless, the WPT successfully charged these two series-connected 12 V, 0.8 Ah lead-acid batteries to their full capacity within a 5-minute charging duration, confirming the charging capability. 

\section{Conclusions and Future Works}
This paper investigated WPT performance for a mid-sized inspection robot, analyzing the impact of various parameters. The paper also presented the implementation of resonant inductive coupling WPT, demonstrating the potential to wirelessly charge mobile robot batteries. The system's efficiency, at 47.14\%, is affected by heat loss and reduced coupling due to coil stretching. The efficiency of the WPT stage is close to the maximum output power point of 50\% where the stresses of passive components have been well utilised in consideration of lightweight Litz wire, with each coil weighing 320 g held together with the octagon-shaped coil frames of 3.54 kg at each side, thereby fully exploiting the stress capabilities of the components. Future work will propose a collapsible coil frame design for enhanced robustness and flexibility.

\end{document}